\RequirePackage{fix-cm}
\documentclass{svjour3}                     % onecolumn (standard format)
\smartqed  % flush right qed marks, e.g. at end of proof
\usepackage{CJKutf8}
\usepackage{footnote}
\usepackage{lmodern}
\usepackage{mdwlist}
\usepackage{url}
\usepackage[hidelinks,bookmarks=false]{hyperref}
\usepackage[official]{eurosym}
\usepackage{epsfig, times}
\usepackage{algpseudocode}
\usepackage{algcompatible}
\usepackage{longtable}
\usepackage{fancyhdr}
\usepackage{amsmath}
\usepackage{float}
\usepackage{verbatim}
\usepackage{afterpage}
\usepackage{hyperref}
\usepackage{tablefootnote}
\usepackage[toc,acronym,xindy]{glossaries}
\usepackage[colorinlistoftodos,prependcaption,textsize=tiny]{todonotes}
\usepackage{color}
\usepackage{amsfonts}
\usepackage{tabularx}
\usepackage[T1]{fontenc}
\usepackage[utf8]{inputenc}
\usepackage{graphicx}
\usepackage{adjustbox}
\usepackage{subfigure}
\usepackage{arabtex}
\usepackage{utf8}
\usepackage{gensymb}

\usepackage{epsfig}
\usepackage{wrapfig}
\usepackage[ruled]{algorithm2e}

\usepackage{multirow}

\begin{document}

\title{A Named Entity Recognition and Topic Modeling-based Solution for Locating and Better Assessment of Natural Disasters in Social Media}
%\subtitle{Disaster event detection, linking and summarization}


\author{\mbox{Ayaz Mehmood}         \and
        \mbox{Muhammad Tayyab Zamir}         \and
        \mbox{Muhammad Asif Ayub}         \and  
	    %\mbox{Talhat Khan}         \and
         \mbox{Nasir Ahmad}         \and  
        \mbox{Kashif Ahmad}
       }


\institute{Ayaz Mehmood \at
              Department of Computer Systems Engineering, University of Engineering and Technology, Peshawar, Pakistan. \\
              \email{ayazmehmood550@gmail.com}           %  \\
           \and
           Muhammad Asif Ayub \at
              Department of Computer Systems Engineering, University of Engineering and Technology, Peshawar, Pakistan. \\
              \email{asifayub836@gmail.com}
              \and 
              Muhammad Tayyab Zamir \at 
              Abbasyn University, Islamabad Campus, Pakistan.
%National University of Science and Technology, Islamabad, Pakistan. \\ \email{gulasma24@gmail.com}
             \and
                Nasir Ahmad \at 
              University of Western Australia, Australia \\ \email{n.ahmad@uetpeshawar.edu.pk}
              \and
                Kashif Ahmad \at 
              Department of Computer Science, Munster Technological University Cork, Ireland \\ \email{kashif.ahmad@mtu.ie}
              \and
              %A. Al-Fuqaha \at
              %Hamad Bin Khalifa University, Doha, Qatar. \\
              %\email{aalfuqaha@hbku.edu.qa }
}


\maketitle

\begin{abstract}
Over the last decade, similar to other application domains, social media content has been proven very effective in disaster informatics. However, due to the unstructured nature of the data, several challenges are associated with disaster analysis in social media content. For example, social media content is generally very noisy containing disaster-related words (e.g., floods, wildfires) used in a different context. Similarly, most of social media posts either do not contain geo-location information or do not represent the actual disaster location. To fully explore the potential of social media content in disaster informatics, access to relevant content and the correct geo-location information is very critical. In this paper, we propose a three-step solution to tackling these challenges. Firstly, the proposed solution aims at the classification of social media posts into relevant and irrelevant posts followed by the automatic extraction of location information from the posts' text itself through Named Entity Recognition (NER) analysis. Finally, to quickly analyze the topics covered in large volumes of social media posts, we perform topic modeling resulting in a list of top keywords, that highlight the issues discussed in the tweet. For the Relevant Classification of Twitter Posts (RCTP), we proposed a merit-based fusion framework combining the capabilities of four different models namely BERT, RoBERTa, Distil BERT, and ALBERT obtaining the highest F1-score of 0.933 on a benchmark dataset. For the Location Extraction from Twitter Text (LETT), we evaluated four models namely BERT, RoBERTa, Distil BERTA, and Electra in an NER framework obtaining the highest F1-score of 0.960. For topic modeling, we used the BERTopic library to discover the hidden topic patterns in the relevant tweets. The experimental results of all the components of the proposed end-to-end solution are very encouraging and hint at the potential of social media content and NLP in disaster management.
\end{abstract}

\section{Introduction}
\label{sec:introduction}
Natural disasters, which refer to hazardous events caused by different geophysical, hydrological, climatological, and meteorological elements, frequently occur in different parts of the world. These disasters generally hurt human lives and infrastructure, resulting in the loss of billions of dollars each year. According to EM-DAT \cite{EM-DAT}, an international disaster database, these disasters also resulted in the loss of 12,000 lives globally in 2023. More than half of these casualties are from low and lower-middle-income countries. Though these disasters can't be prevented, there are ways to minimize their adverse impact on human lives and infrastructure. For instance, the impact of disasters could be minimized through proper planning by allocating and supplying adequate medical supplies, shelters, food, and other resources to the affected areas. 

However, for better planning and resource allocation, access to relevant and timely information (e.g., the scale of the damage to the infrastructure, the number of affected people, the needs of the affected people, etc.,) is very critical. The literature reports several situations where access to relevant information may not be possible due to different reasons, such as the unavailability of reporters in the area and damage to communication infrastructure \cite{said2019natural}. To overcome these challenges, over the last decade, social media platforms, such as Twitter, Facebook, and Instagram, have been widely explored as a potential source of communication during natural disasters \cite{alam2023role}. Despite the great success, several challenges are associated with the extraction of meaningful information. For instance, social media content is generally very noisy containing a lot of irrelevant content tagged with the relevant keywords. More importantly, the majority of
social media posts either do not contain geo-location information or do not represent the
actual disaster location thus making it difficult to locate the disaster-affected areas \cite{ahmad2019automatic}. Similarly, due to the large volume of information, it is very challenging to manually analyze the content of the relevant posts.    

%%Considering the importance and applications of social media content in disaster analytics floods detection in social media content has been also included in the MediaEval benchmark competition as a shared task for several years. This paper presents a solution for the MMDisaster task presented in MediaEval 2022 \cite{andreadis2022disastermm}. The challenge aims to solve two key challenges to disaster analytics in social media. The first subtask aims at reducing social media noise by automatically filtering social media content to obtain relevant content. The second subtask aims at extracting location information from social media text, allowing automatic positioning of a potential incident due to floods. For both subtasks, we proposed several interesting solutions as described in Section \ref{sec:approach}.
To overcome these challenges, in this paper, we propose a three-step Natural Language Processing (NLP)-based solution able to automatically differentiate between relevant and irrelevant content, extract location information from the posts' text itself, and list the key topics discussed in the social media posts relevant to disaster affected areas. For the first task (i.e., text classification), we propose a merit-based fusion framework incorporating the text analysis capabilities of several transformer-based NLP models. For the second task (i.e., identification and extraction of location/places information from text), we propose a Named Entity Recognition (NER) framework able to accurately identify and extract different addresses/places mentioned in the relevant social media posts. Finally, in the third step, we perform topic modeling using the BERTopic library \cite{grootendorst2022bertopic} to discover the hidden topic patterns in the relevant social media posts.

The key contributions of the work can be summarized as follows:

\begin{itemize}
    \item We propose an end-to-end solution tackling three key challenges associated with disaster informatics in social media content including filtering out irrelevant content, extracting the location of affected areas, and automatically identifying topics/keywords discussed in the relevant posts. 
    \item Firstly, we propose a merit-based fusion framework by combining the classification capabilities of several state-of-the-art NLP models to differentiate between relevant and irrelevant social media posts.
    \item We also propose a transformer-based NER framework to automatically identify and extract addresses/places mentioned in social media posts.
    \item We also propose a BERTopic-based topic modeling to automatically analyze and extract key topics discussed in social media posts relevant to a particular location.
\end{itemize}

The rest of the paper is organized as follows. Section \ref{sec:related_work} provides an overview of the related literature. Section \ref{sec:tasks} explains the objectives of each task while Section \ref{sec:methodology} describes the methodology of the proposed solution. Section \ref{sec:dataset} and Section \ref{sec:experiments} discuss the dataset and experimental results, respectively. Finally, Section \ref{sec:conclusion} concludes the paper.

\section{Related Work}
\label{sec:related_work}
The literature already demonstrates the effectiveness of social media content in different application domains for different tasks, such as social events detection \cite{ahmad2019deep}, tourism promotion \cite{nusair2024exploring}, public health \cite{batra2023effective}, education \cite{ahmad2023data}, and feedback on public services and infrastructure, such air quality and water networks \cite{ahmad2022social}. Social media content has also been widely explored for different tasks during natural disasters \cite{said2019natural}. For instance, Ahmad et al. \cite{ahmad2017jord} proposed an end-to-end system able to detect disaster events in social media and automatically collect relevant images, videos, and text, providing a detailed overview of the events. The system also utilizes satellite imagery to provide a better view of the disaster-affected areas. Similarly, Yap et al. \cite{yap2023use} explored the potential of social media as a source of communication during natural disasters. Zhang et al. \cite{zhang2020hybrid} analyzed unlabelled social media posts including text and visual content for damage assessment in disaster-affected areas. The authors proposed a hybrid transfer learning framework enabling effective damage severity assessment of recent disaster events based on earlier events by employing a pre-trained disaster damage severity deep model. Ahmad et al. \cite{ahmad2019automatic} explored another interesting application of disaster informatics by analyzing satellite imagery for the identification of passable and non-passable roads during floods. Social media content has also been used for the identification and classification of the needs of affected individuals \cite{alam2020deep} and for monitoring rescue and reconstruction activities \cite{jamali2019social}. More recently, Hassan et al. \cite{hassan2019sentiment,hassan2022visual} analyzed the perception and emotions of viewers evoked by natural disaster-related social media imagery. The authors highlighted how this application of disaster informatics could help news agencies, public authorities, and aid agencies with different tasks.  

Despite the great success and impact on different applications, the extraction of meaningful insights from social media content is very challenging in several ways. For example, social media outlets generally contain a lot of noisy, irrelevant, or less informative content, and require extensive filtering. The literature already reports several interesting solutions for content-based filtering of social media content, allowing the extraction of disaster-related posts from large volumes of social media posts \cite{said2019natural}. For instance, Ahmad et al. \cite{ahmad2018comparative} proposed a deep and handcrafted features-based framework for the identification and classification of disaster-related social media data. Another key challenge associated with social media-based disaster informatics is the unavailability and accuracy of geo-location information associated with social media \cite{alam2023role}. To overcome this challenge, the literature reports some initial efforts employing interesting strategies. For instance, Sathianaray et al. \cite{sathianarayanan2024extracting} analyzed and extracted disaster locations from relevant social media imagery using a deep learning-based image processing framework. To this aim, the authors utilized the phone numbers available in social media imagery. However, the availability of images with phone numbers in disaster-related social media posts is not guaranteed. Some works also explored the possibility of extraction of location information from social media text \cite{andreadis2022disastermm,suleman2023floods}. Another key challenge associated with disaster analysis in social media is manually analyzing and extracting topics covered in the large volumes of social media posts. 

In this work, we explore three key challenges associated with disaster informatics in social media by proposing a three-step complete framework with text classification, NER, and topic modeling components.

\section{Objectives}
\label{sec:tasks}
The proposed framework aims to address the three key challenges in disaster informatics including (i) the differentiation between relevant and irrelevant tweets, (ii) the extraction of geo-location information of the affected areas, and (iii) automatically extracting the topics/issues discussed in the relevant posts. A brief overview of each of the tasks and associated challenges is provided below.

\begin{itemize}
    \item \textbf{Relevance Classification of Twitter Posts (RCTP):} One of the key challenges associated with social media content is dealing with the large volumes of noisy data containing posts that do not refer to actual cases of disasters but rather simply contain disaster-related keywords. Thus, crawling social media platforms using disaster-related keywords may result in irrelevant content. This challenge could be overcome with content-based filtering by analyzing the content of each tweet tagged with disaster-related keywords. Manually analyzing each post is a tedious and time-consuming process. An NLP-based automatic analysis of social media content allows to filter out such irrelevant posts. 

\item \textbf{Location Extraction from Twitter Texts (LETT):} The other key challenge associated with social media-based disaster informatics is the identification of the affected areas due to the unavailability of the geolocation information. The majority of social media posts are either not geotagged or their geoinformation is questionable. There are several reasons for missing or wrong geolocation information associated with social media posts. For example, the geolocation information is not available if the users turn off the location features/services on their devices while posting. Moreover, the geolocation associated with social media posts refers to the location from where the tweets are posted and it may not be necessary that the users post the disaster information from the affected areas. One of the potential solutions to this challenge is the extraction of location information from the Twitter text. It is very common to find location information in the social media text. In this case, a NER framework could be used to automatically identify and extract the location information/addresses in large volumes of social media text.

\item \textbf{Topic Modeling for the extraction of Key Topics:}  Analyzing the large volumes of relevant information available on social media is also very challenging and time-consuming. However, thanks to new developments in NLP, it is possible to quickly and automatically discover hidden topical patterns in large volumes of data, which could be then used as meaningful insights for data-driven decisions. Thus, through topic modeling, we can automatically identify and extract key topics discussed in the relevant tweets. 
 
\end{itemize}

\section{Methodology}
\label{sec:methodology}
\begin{figure}[]
%\label{fig:taxonomy}
\centering
\includegraphics[width=0.70\textwidth]{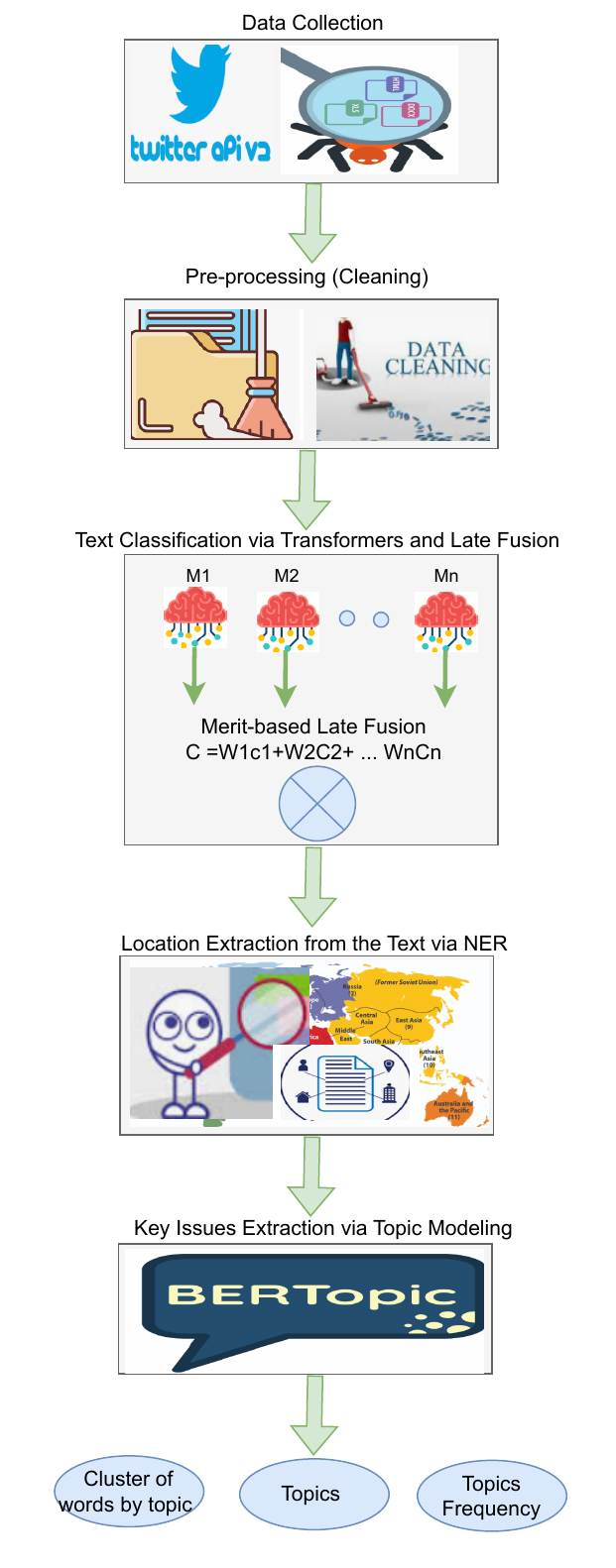}
\caption{A block diagram of the proposed methodology.}
	\label{fig:methodology}
\end{figure}

Figure \ref{fig:methodology} provides a block of the proposed solution. The proposed solution is composed of five components including data collection, pre-processing, classification of the tweets into relevant and irrelevant, location information extraction through NER, and topic modeling to automatically identify and extract the topics discussed in the relevant tweets. In the pre-processing step, we cleaned the dataset by removing less informative/unnecessary text, such as URLs, stop and short words, etc., For text classification, we proposed a merit-based late fusion framework incorporating the capabilities of several transformers-based NLP models by assigning them weights based on their performance. Similarly, we employed several transformers-based NLP models in an NER framework for the identification and extraction of the location information from the text of the relevant tweets. For topic modeling, we rely on an open-sourced library namely BERTopic to identify hidden topics in the relevant tweets. A detailed description of each step is provided below.
%%%%%%%%%%%%%%%%%%%%%%%%%%%%%%%%%%%%%%%%%
%%%%%%%%%%%%%%%%%%%%%%%%%%%%%%%%%
\subsection{Relevance Classification of Twitter Posts (RCTP)}
For this task, we explored different options by starting with analyzing the quality of the available data including the Twitter text and associated images and meta-data. During our analysis, we noticed that the majority of Tweets do not contain any visual content (images and videos). Moreover, the majority of the images associated with the tweets were irrelevant. Thus, we decided to use textual information only in our solution. Our text classification framework is composed of three components including (i) text pre-processing, (ii) text classification through different state-of-the-art NLP models, and (iii) fusion of the classification scores obtained through the individual models. During pre-processing, we cleaned the data by removing unnecessary information, such as usernames, URLs, emojis, and stop words. %Figure \ref{fig:frequency} shows the list of the most frequently occurring words in the cleaned dataset. 
%%%%%%%%%%%%%%%%%%%%%%%%%%%55
%\begin{figure*}[t]
%\centering
% \resizebox{0.99\columnwidth}{!}{
%\includegraphics[width=0.72\textwidth]%{words_frequency.jpg}
% }
% \caption{Frequencies of the most frequently occurring words in the cleaned dataset.}
%\label{fig:frequency}
%\vspace{-0.45cm}
%\end{figure*}
%%%%%%%%%%%%%%%%%%%%%%%%%%%
After pre-processing, several state-of-the-art NLP algorithms including BERT \cite{devlin2018bert}, Roberta \cite{liu2019roberta}, Distil BERT \cite{sanh2019distilbert}, and ALBERT \cite{lan2019albert} are fine-tuned on the training set. During finetuing our models, we Adam optimizer with a binary cross-entropy cost function. Moreover, we used a mini-batch size of 32 for 20 epochs. 

The final phase of the proposed framework for the text classification part is based on late fusion, where we employed several weight optimization/selection methods to assign merit-based weights to the models using \ref{eq:fusion}. In the equation, $S_{c}$ represents the resultant final joint/combined classification score while $w_{n}$ is the weight assigned to the nth model having individual classification score/probability of $C_{n}$. Moreover, ${n}$ represents the total number of ML models employed in this work.
%%%%%%%%%%%%%%%%%%%
\begin{equation}
S_{c}=W_{1}C_{1}+W_{2}C_{2}+... +W_{n}C_{n}  
\label{eq:fusion}
\end{equation} 
%%%%%%%%%%%%%%%%%%%%%%

The weights assigned to the models (i.e., $W_{1}=W_{2}=W_{n}=1 = 1/N$) are selected through different optimization/search methods. An overview of the weights optimization/selection methods is provided below.

\begin{itemize}
    \item \textbf{Particle Swarm Optimization:}
The selection of the method for weight optimization in this work is based on its proven performances in similar tasks \cite{ahmad2020intelligent,ahmad2018ensemble}. PSO is a heuristic approach and provides a near-optimal solution for optimization problems. Apart from being easy to understand and implement, PSO brings several other advantages over other optimization methods. For instance, it is efficient and insensitive to scaling and requires fewer parameters, which makes it a suitable choice for concurrent parameters. On the other hand, it also has several limitations, such as a low convergence rate and may stuck in local minima, etc., 

PSO starts with an initial solution (i.e., a set of weights) and improves it iteratively by evaluating it using the criteria provided in its fitness/objective function.  In this work, each set/combination of weights is the potential solution. Moreover, our objective function is based on the accumulative error ($e$), which is computed using Equ. \ref{fitness_function}.
%%%%%%%%%%%%%%%%%%%%%%%%%%%%%%%%%%%%%%%%%%%%%%
\begin{equation}
e = 1-A_{acc}
	\label{fitness_function}
\end{equation}
%%%%%%%%%%%%%%%%%%%%%%%%%%%%%%%%%%%%%%%%%%%%%
In the equation, $e$ and $A_{acc}$ represent the error and the cumulative accuracy, which are computed on a separate validation set using Equ. \ref{accuracy_function}. Here $p_{n}$ denoted the posterior probability obtained with the $n^{th}$ model while $x(n)$ is the weight to be assigned to the model.
  %%%%%%%%%%%%%%%%
\begin{equation}
%\small
%\centering
A_{acc} = x(1)*p_{1}+x(2)*p_{2}+... +x(n)*p_{n} 
\label{accuracy_function}
\end{equation}
%%%%%%%%%%%%%%%%%%%%%%%%%%%%%%%%%%%%%%%%
\item \textbf{Nelder Mead Method:}
Our second fusion method is based on the Nelder-Mead algorithm, which is suitable for both one-dimensional and multi-dimensional optimization problems \cite{singer2009nelder}. However, it also has some limitations. For example, it is not efficient and generally takes a large number of iterations to find a suitable solution. This method generates $n+1$ dimensions for a $n$ dimensional task and computes the values for the objective function at each point, iteratively. In this work, the number of dimensions is 3 (i.e., $n=3$) while our objective function is defined by Equ. \ref{fitness_function}.

   \item \textbf{Powell Method:}
Our third fusion method is also based on an evolutionary method namely Powell's, which is a modified version of the original algorithm \cite{powell1964efficient}, by introducing a stochastic element. Similar to other evolutionary methods, the algorithm starts with an initial solution (i.e., a set of parameter values) and improves it iteratively in several rounds. Each solution is checked against the criteria provided in the fitness function in every iteration/generation. The algorithm moves in a single direction until a local minima is found. In this work, the objective function is based on the cumulative error as defined in Equ. \ref{fitness_function}.
\end{itemize}

\subsection{Location Extraction from Twitter Texts (LETT)}
The LETT task is tacked with a transformers-based  NER framework by locating and classifying the locations mentioned in the Twitter text. NER is one of the key tasks in NLP and involves extracting and differentiating items (named entities) from other items having similar attributes, such as names of persons/places, contact details, and addresses/geo-location. In this work, we are interested in extracting words representing the starting and subsequent words in text referring to a location mentioned in the Twitter posts. To this aim, we employed several state-of-the-art LLMS including BERT, Roberta, Distil BERT, and ELECTRA.  ELECTRA is a new pretraining approach that uses two different transformer-based models, namely the (i) generator and the (ii) discriminator. All these models are fine-tuned for the task on the new dataset. During fine-tuning the models, we used the same hyper-meter settings. We used the Adam optimizer with a batch size of 16 and a learning rate of 0.00001 for 5 epochs. Moreover, since annotations are provided at the word level, no pre-processing techniques are used, rather the models are trained on the raw data. 

\subsection{Topic Modeling}
The last component of the proposed framework is based on topic modeling, another interesting NLP application. Topic modeling allows identification and grouping of similar words from text using unsupervised ML techniques. This component enables the proposed tool to automatically analyze the large collection of tweets and produce a list of topics covered in the relevant tweets by discovering the hidden topical patterns. This automatic analysis and extraction of key topics from the relevant tweets reduces the efforts and time required in manual analysis of the large volumes of relevant tweets for data-driven decisions.

For topic modeling, we used the BERTopic \cite{grootendorst2022bertopic}, an open-source library. As a first step, we performed some pre-processing on the tweets by removing stop words, URLs, and short words, having 3 or fewer 3 characters. Subsequently, the tweets are converted into embeddings through a pre-trained model called SentenceBERT. This step is followed by a Uniform Manifold Approximation and Projection(UMAP)-based dimensionality reduction, which also involves the selection of several parameters. The key parameters fine-tuned in this work include "n\_neighbours" and "n\_components", which are set at 15 and 5, respectively. The parameter ''n\_neighbours'' represents the neighboring samples for manifold approximation, and a larger value generally results in large clusters. On the other hand, ''n\_components'' is the reduced dimensionality of the embeddings and is, typically, kept very low.  The dimensionality reduction phase is followed by HDBSCAN (Hierarchical DBSCAN)-based clustering. In the final step, topics are extracted from the clusters c-TF-IDF algorithm, which is an extension of TF-IDF (Term Frequency-Inverse Document
Frequency). c-TF-IDF is adjusted to work well with clusters instead of documents. After extracting topics, the posterior probabilities of the topic prediction for each tweet are provided to highlight the predictions' confidence.

In topic modeling, we conducted several types of experiments, which are described in \ref{sec:experiments} in detail. 

%The algorithm used in this work extracts topics from Tweets in three different steps starting from converting the tweets into embeddings, then reducing the dimensionality and clustering, and finally converting them into topics. The embeddings are obtained by a pre-trained model namely Sentence-BERT. The dimensionality reduction and clustering are carried out through Uniform Manifold Approximation and Projection(UMAP) and HDBSCAN (Hierarchical DBSCAN), respectively. Finally, topics are extracted from the clustering using a modified form of TF-IDF (Term Frequency-Inverse Document Frequency) namely c-TF-IDF.

%The algorithm brings several advantages. For instance, it clusters documents based on both lexical and semantic similarities. Moreover, BERTopic provides a library with several packages allowing more accurate and better visualization of the clusters, topics, and probabilities. It also comes with a few limitations. For instance, its assumption that each document/tweet contains only one topic is its main limitation, though it is possible to have Tweets with multiple topics. We note that we also performed some pre-processing in addition to the data cleaning before topic modeling. For instance, we removed short and stop words, numbers, and alphanumeric characters. This allows us to remove irrelevant frequently used words.  

\section{Dataset}
\label{sec:dataset}
The dataset used for the validation of the proposed solutions is released in a benchmark competition namely ''DisasterMM: Multimedia Analysis of Disaster-Related Social Media Data'' \cite{andreadis2022disastermm}. The dataset is released in two different parts: one for the text classification task (i.e., RCTP) and the other for location extraction from Twitter texts (LETT). The RCTP dataset is composed of 8,000 tweets, which were collected between May 25, 2020, and June 12, 2020. The tweets were downloaded using flood-related keywords in the Italian Language, such as ''alluvione'',  ''allagamento'', and ''esondazione''. The collected data include tweet text, images, and meta-data. The data samples are annotated with either 1 (relevant) or 0 (not relevant). Moreover, the dataset is provided in two different sets namely the development set and the test set. The development set is composed of 5,337 tweets while the test set contains a total of 1,315 tweets.  

On the other hand, the LETT dataset contains 6,000 tweets collected between March 25, 2017, and August 1, 2018. Similar to RCTP, the dataset is collected using flood-related  Italian keywords. Moreover, the dataset is annotated per word where each word is annotated with either ''B-LOC'', ''I-LOC'', or ''O''.  ''B-LOC'' and ''I-LOC'' represent the first and the subsequent words of a sequence referring to a location, respectively while the non-location word is annotated with ''0''. For instance, the labels for each word in the sentence ''Allagamento in via Prati della Farnesina'' are ''O O B-LOC I-LOC I-LOC I-LOC''. 

\section{Experiments}
\label{sec:experiments}

\subsection{Experimental Results of the Text Classification Task}
Table \ref{tab:RCTP} provides the experimental results of the text classification task for the individual models and fusion methods. In terms of individual performances, overall better results are obtained for all the models with RoBERTA, BERT, and GPT producing the highest F1-score of 0.924. As far as the fusion of the models is concerned, the results are slightly improved compared to the best-performing individual models. One of the potential reasons for the less improvement with the fusion could be the high performances of the individual models themselves and probably there's less room for improvement. Among the fusion methods, the Powell method proved to be more effective in terms of improving the F1 score. Overall the results of the task are very encouraging, showing the potential NLP techniques in the domain. 

%%%%%%% Experimental Results %%%%%%%%%%%%%%%%%%%%%%%%%%
% \usepackage{multirow}
\begin{table}[]
\centering
\caption{Experimental results of the RCTP task.}
		\label{tab:RCTP} 
\begin{tabular}{|c|c|c|c|}
\hline
\textbf{Method} & \textbf{F1 Score} & \textbf{Method} & \textbf{F1 Score} \\ \hline
 Multi.lig-BERT-uncased &	0.9244 & Albert	& 0.9096  \\ \hline
 Distillbert	& 0.9218  & XLM - Roberta-base	& 0.9244  \\ \hline
 GPT	& 0.921  &  simple fusion	& 0.9287  \\ \hline
 PSO Method	& 0.9265 &  Nelder Mead Method	& 0.9244  \\ \hline
   Powell Method	& 0.9331 & - &	-  \\ \hline
   % trust-constr &	0.9331 & -  & - \\ \hline
\end{tabular}
\end{table}
%%%%%%%%%%%%%%%%%%%%%%%%%%%%%%%%%%%%%%%%%%%%%%
Table \ref{tab:Comparision_RCTP} provides the proposed solution against comparisons against existing methods on the RCTP task. All these existing methods are proposed in a recent benchmark competition on the task namely DisasterMM in MediaEval 2022 \cite{andreadis2022disastermm}. Overall, the F1 scores of the majority of methods on the task are very high, indicating the effectiveness of NLP techniques in the application. Our solution based on the Powell method is among the best-performing methods.  Our best-performing solution produced the highest F1-score of 0.933, beating the existing best-performing method with more than 1.3\%.  

%%%%%%%%%%%%%%%%%
\begin{table}[]
\centering
\caption{Comparison against the existing methods on RCTP dataset.}
		\label{tab:Comparision_RCTP} 
\begin{tabular}{|c|c|}
\hline
\textbf{Method} & \textbf{F1 Score} \\ \hline
Christodoulou et al. \cite{christodoulou2022model} (textual features) &  0.920\\ \hline
Christodoulou et al. \cite{christodoulou2022model} (textual and visual features)  & 0.853 \\ \hline
Suleman et al. \cite{suleman2023floods} & 0.792 \\ \hline
 Delacruz et al. \cite{dela2022understanding} &  0.742\\ \hline
  Shao et al. \cite{shao2022hffd} & 0.910 \\ \hline
Mukhtar et al. \cite{mukhtiar2023relevance}   & 0.878  \\ \hline
 This work    & 0.933 \\ \hline
\end{tabular}
\end{table}
%%%%%%%%%%%%%%%%

\subsection{Location Identification via NER}
Table \ref{tab:NER} presents the experimental results of the proposed framework on the LETT task. To align it with the MediaEvals DisasterMM benchmark task \cite{andreadis2023disastermm}, two different types of F1 scores are reported. These scores include \textit{Exact F1-Score} and \textit{Partial F1-Score}. For the exact F1-score, the predicted labels have to fully match with the ground truth while in the case of partial F1-score either “B-LOC” or “I-LOC” are considered true labels as long as the annotator’s label concerns location. Similar to RCT, overall the results are very encouraging. Overall better results are obtained for the proposed framework with BERT and Electra models by F1-scores of 0.962 and 0.9607, respectively. These results indicate the suitability of the proposed solution for tackling this challenging and crucial task of social media analytics.

%%%%%%% Experimental Results %%%%%%%%%%%%%%%%%%
% \usepackage{multirow}
\begin{table}[]
\centering
\caption{Experimental results of the NER task.}
		\label{tab:NER} 
\begin{tabular}{|c|cc|}
\hline
\multirow{2}{*}{\textbf{Method}} & \multicolumn{2}{c|}{\textbf{F1-score}} \\ \cline{2-3} 
 & \multicolumn{1}{c|}{\textbf{Partial}} & \textbf{Pre-Exact} \\ \hline
 Roberta 	& \multicolumn{1}{c|}{0.9311} &  0.9359 \\ \hline
 Distillbert& \multicolumn{1}{c|}{0.9356} & 0.9444 \\ \hline
 BERT& \multicolumn{1}{c|}{0.9514} & 0.962 \\ \hline
Electra & \multicolumn{1}{c|}{0.9491} & 0.9607 \\ \hline
 %& \multicolumn{1}{c|}{} &  \\ \hline
\end{tabular}
\end{table}

%%%%%%%%%%%%%%%%%%%%%%%%%%%%%%%%%%%%%%%%%%%%%%
We also provide a comparison against the existing methods presented in the benchmark competition for the task. In total, two methods were proposed for the task in the competition. As can be seen in Table \ref{tab:comparison_NER}, the proposed solution provides significant improvements in terms of both exact and partial F1 scores.

%%%%%%% Experimental Results %%%%%%%%%%%%%%%%%%
% \usepackage{multirow}
\begin{table}[]
\centering
\caption{Comparison against the existing methods on the NER task.}
		\label{tab:comparison_NER} 
\begin{tabular}{|c|cc|}
\hline
\multirow{2}{*}{\textbf{Method}} & \multicolumn{2}{c|}{\textbf{F1-score}} \\ \cline{2-3} 
 & \multicolumn{1}{c|}{\textbf{Partial}} & \textbf{Pre-Exact} \\ \hline
Delacruz et al. \cite{dela2022understanding} 	& \multicolumn{1}{c|}{0.834} &  0.817 \\ \hline
 Sulemain et al. \cite{suleman2023floods} & \multicolumn{1}{c|}{0.671} & 0.60 \\ \hline
 This Work (BERT-based solution)& \multicolumn{1}{c|}{0.9514} & 0.962 \\ \hline
This Work (Electra-based solution) & \multicolumn{1}{c|}{0.9491} & 0.9607 \\ \hline
 %& \multicolumn{1}{c|}{} &  \\ \hline
\end{tabular}
\end{table}

%%%%%%%%%%%%%%%%%%%%%%%%%%%%%%%%%%%%%%%%%%%%%%
\subsection{Topic Modeling}
In this section, we provide the experimental results of the topic modeling part. The objectives of this experiment are manifold. Firstly, we want to extract the topics discussed in all the relevant tweets. Secondly, we are interested in more localized topics (i.e., the topics discussed in tweets relevant to a particular location). We are also interested in topics relevant to mostly discussed locations in the relevant tweets.  

To achieve these objectives, we performed the following experiments.
\begin{itemize}
    \item Topic modeling of all the relevant tweets to discover the topics discussed in the disaster-related tweets. This will give us an idea of the most commonly discussed topics in disaster-related tweets.
    \item Topic modeling on tweets relevant to most frequently discussed locations in the tweets. This experiment allows us to analyze any correlation in topics discussed in the most frequently discussed locations. The location information is obtained from the NER part. 
    \item Topic modeling of the tweets relevant to each single location individually. This allows us to analyze more localized issues and needs of the affected areas. 
\end{itemize}

\subsubsection{Topic Modelling of all Tweets}
\begin{figure}[]
%\label{fig:taxonomy}
\centering
\includegraphics[width=0.99\textwidth]{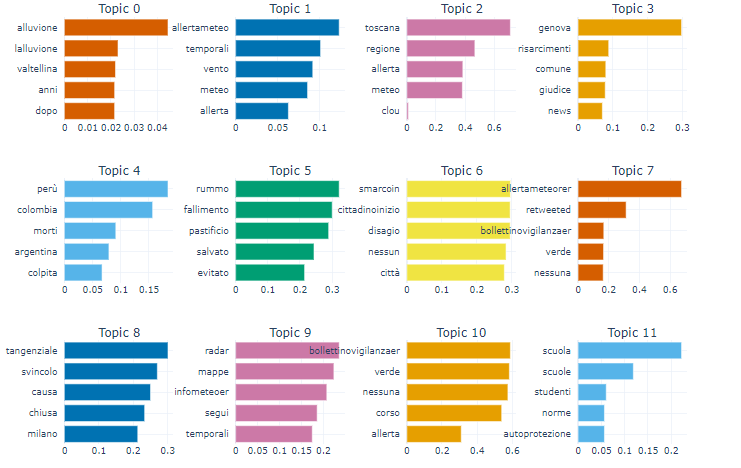}
\caption{Word scores of the topics extracted from the complete set of relevant tweets.}
	\label{fig:complete-topics}
\end{figure}

Figure \ref{fig:complete-topics} provides a summary of the top 12 topics, along with the word scores, extracted from the complete set of relevant tweets. The length of the bars shows the score of each keyword relevant to the topic, thus, a longer bar means high importance/relevance of the keyword to the topic. For instance, the keywords ''alluvione'' and ''llaluvione'' are the most important/contributing keywords to topic 0. ''Alluvione'' and ''llaluvione'' are Italian words that mean ''Flood'' in English. Furthermore, the third most contributing/important keyword to topic 0 is ''Valtellina'', a valley in Itlay. Thus, this indicates that the tweets belonging to topic 0 are relevant to floods in ''Valtellina'' and may provide useful information and thus need further analysis. Similarly, the keywords, such as ''allartameteo (weather alert)'', ''vento (wind)'' and ''meteo (weather) '' in topic 1 indicate weather alerts due to strong winds. We also automatically classify the tweets based on the topics, which allows the extraction of such relevant and important tweets. 

%%%%%%%%%%%%%%%%%%%%%%%%%%%%%%%%%%%%%%%%%
%%%%%%%%%%%%%%%%%%%%%%%%%%%%%%%%%

\subsubsection{Topic Modelling of Tweets relevant to frequently mentioned Locations}
For this experiment, first, we automatically identified the top 10 frequently mentioned locations. These locations include Valtellina, Smarcoin, Toscana, Genova, Benevento, Firenze, Sardegna, Sangiorgiown, and Italia.

\begin{figure}[]
%\label{fig:taxonomy}
\centering
\includegraphics[width=0.99\textwidth]{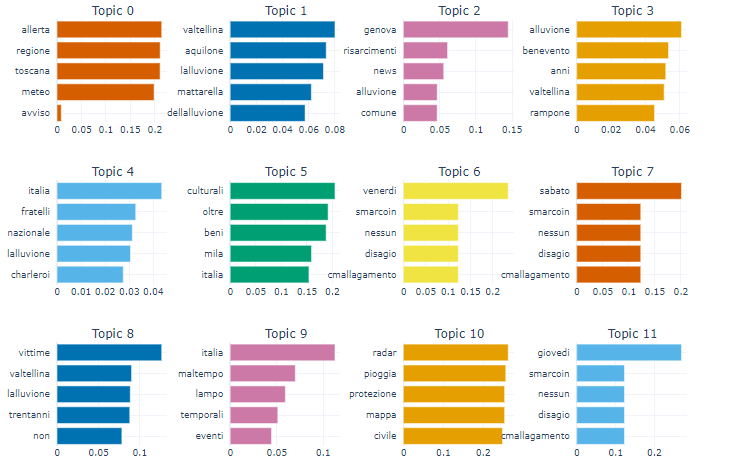}
\caption{Word scores of the topics extracted from the tweets relevant referring to frequently mentioned location.}
	\label{fig:frequent-topics}
\end{figure}

Figure \ref{fig:frequent-topics} provides a list and scores of relevant keywords for the first 12 topics discussed in tweets referring to frequently mentioned locations. These keywords provide interesting insights. For example, the keywords in topic 0 include ''allerta (alert)'', ''region'', ''Tuscana (a regions in Italy)'', ''meteo (weather)'', and ''Avvisio (notify)''. The combination of these keywords reveals that the tweets belonging to this topic mostly provide alerts about bad weather in the Tuscana region. Similarly, topics 1, 2, and 3 refer to flood-related tweets in other regions, such as Genevo and  Valtellina. These results motivated us for the three experiments where we performed topic modeling on tweets referring to specific regions.
%%%%%%%%%%%%%%%%%%%%%%%%%%%%%%%%%%%%%%%%%
%%%%%%%%%%%%%%%%%%%%%%%%%%%%%%%%%

\subsubsection{Location-based Topic Modelling}
\begin{figure}[!ht]
  \centering
\includegraphics[width=.4\textwidth]{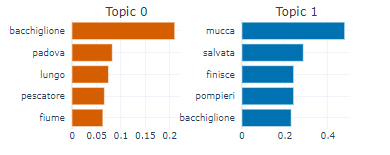}\quad
  \includegraphics[width=.4\textwidth]{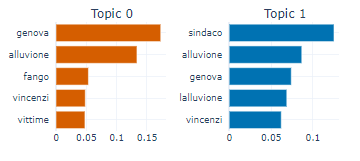}\\
  \includegraphics[width=.4\textwidth]{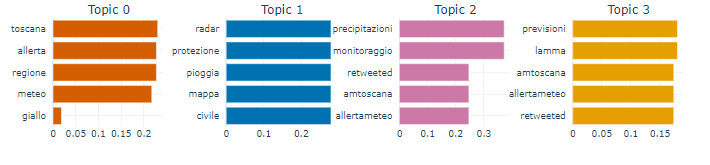}\quad
  \includegraphics[width=.4\textwidth]{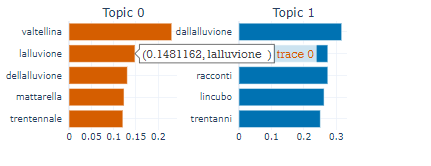}
  \caption{Topic modeling of tweets belonging to specific locations. Top left: Bacchiglione, top right: Genova, bottom left: Toscana, bottom right: Valtellina.}
  \label{fig:local_regions}
\end{figure}

In this experiment, we analyze the topics extracted from individual locations/regions. We are reporting the results of some of the regions as proof of concept to show the effectiveness of localized topic modeling. Figure \ref{fig:local_regions} summarizes the keyword scores for topics extracted from tweets belonging to four locations including Bacchiglione, Genova, Toscana, and Valtellina. We note that we kept the maximum number of topics the same for all locations, however, as can be seen in the figure, only two topics are extracted for most of the locations covered in the figure. For instance, only two topics are extracted from the tweets referring to Genova, which are relevant to floods as apparent from the keywords, such as ''alluvione''. Similarly, the other keywords contributing to the topic include ''vittime (victims)'' and ''fungo (fungus)'' indicate that the tweets belonging to this topic contain information about the victims and some diseases. Similarly, topic 1 extracted from the tweets relevant to Bacchiglione include keywords, such as ''bacchiglione'', ''pompieri (firefighters)'', ''salvata (saved)'', and ''mucca (cow)'' indicating the rescue of a cow by firefighters in bacchiglione river.

%%%%%%%%%%%%%%%%%%%%
%%%%%%%%%%%%%%%%%%

\section{Lessons Learned}
The experimental results of the components (i.e., filtering through text classification, location identification through NER, and topic modeling) are very encouraging. These results lead to several interesting insights. Some of the key lessons learned during the experiments are summarized as follows. 

\begin{itemize}
    \item As apparent from the tweets analyzed during the experiments, social media plays a crucial role in disaster analytics by providing more localized and up-to-date information on different aspects, such as the severity of the disasters, damage to properties, and rescue operations. 
    \item Text classification through state-of-the-art NLP algorithms is an effective way of automatically analyzing and filtering out irrelevant content from large volumes of social media data. The performance of the individual LLMs deployed in this work was very encouraging. The fusion of these models further improved the results.
    \item As discussed earlier, the majority of social media posts either do not contain geolocation information or do not represent the actual disaster location thus making it difficult to locate the disaster-affected areas. However, it has been noticed that generally, social media users identify the locations of disaster-affected areas in their posts. NER provides a promising solution to automatically extract locations and addresses mentioned in social media text itself. The performance of all the LLMs deployed for this task is highly encouraging. 
    \item Topic modeling brings several advantages to disaster informatics. It allows the automatic identification of hidden topics in large volumes of social media posts and then categorizes the social media posts into these topics. This avoids the manual analysis of social media posts, which is a tedious and time-consuming task. We found all the topics and associated keywords extracted through BERTopic relevant and meaningful. The combination of these keywords provides enough details to get an idea of the content covered in the social media posts as explained above.   
\end{itemize}

\section{Conclusions}
\label{sec:conclusion}
In this work, we proposed an end-to-end novel framework for disaster informatics. The proposed framework is composed of three components, including text classification, NER for location identification in text, and topic modeling to discover hidden patterns in social media posts. All the components of the proposed framework are evaluated on a benchmark dataset. We found all the solutions encouraging, which could lead to a better assessment of natural disasters. In the current implementation, we analyzed text only. In the future, we aim to explore the potential of video content for similar tasks.

\section{Funding and/or Conflicts of interests/Competing interests}
The authors declare no conflict of interest. 
%\begin{acknowledgements}
%The authors declare no conflict of interest
%\end{acknowledgements}

\bibliographystyle{spmpsci}      % mathematics and physical sciences

\bibliography{sigproc}

\end{document}